\pdfoutput=1

\documentclass[11pt]{article}

\usepackage[]{ACL2023}

\usepackage{times}
\usepackage{latexsym}

\usepackage[T1]{fontenc}

\usepackage[utf8]{inputenc}

\usepackage{microtype}

\usepackage{inconsolata}

\usepackage{amsmath}
\usepackage{graphicx}
\usepackage{booktabs}
\usepackage{arydshln}
\usepackage{tabularx}
\usepackage{amsfonts}
\usepackage{multirow}
\usepackage{multicol}
\usepackage{cleveref}
\usepackage{xspace}
\usepackage{makecell}
\usepackage{fontawesome5}
\usepackage{colortbl}
\usepackage[ruled, lined, linesnumbered, commentsnumbered]{algorithm2e}
\usepackage{todonotes}

\crefformat{section}{\S#2#1#3} %
\crefformat{subsection}{\S#2#1#3}
\crefformat{subsubsection}{\S#2#1#3}
\newcommand{\refequ}[1]{Equation~(\ref{#1})}
\newcommand{\reffig}[1]{Figure~\ref{#1}}
\newcommand{\reftab}[1]{Table~\ref{#1}}

\def\eg{\textit{e.g.}\xspace}

\def\etc{\textit{etc.}\xspace}
\def\ie{\textit{i.e.}\xspace}

\definecolor{myyellow}{RGB}{249, 231, 173}
\definecolor{mygreen}{RGB}{233, 242, 230}

\newcommand{\RNum}[1]{\uppercase\expandafter{\romannumeral #1\relax}}

\DeclareMathOperator*{\argmax}{argmax}

\title{Class-Incremental Learning based on Label Generation}

\author{
Yijia Shao$^{1}$, Yiduo Guo$^{1}$, Dongyan Zhao$^{1,2,3}$~~and Bing Liu$^{4}$\\ 
$^1$Wangxuan Institute of Computer Technology, Peking University\\
$^2$National Key Laboratory of General Artificial Intelligence\quad$^3$BIGAI, Beijing, China\\
$^4$Department of Computer Science, University of Illinois at Chicago\\
\texttt{shaoyj@pku.edu.cn, yiduo@stu.pku.edu.cn, zhaody@pku.edu.cn, liub@uic.edu}\\
}

\begin{document}
\maketitle
\begin{abstract}
Despite the great success of pre-trained language models, it is still a challenge to use these models for continual learning, especially for the \textit{class-incremental learning} (CIL) setting due to \textit{catastrophic forgetting} (CF). This paper reports our finding that if we formulate CIL as a \textit{continual label generation} problem, CF is drastically reduced and the generalizable representations of pre-trained models can be better retained. We thus propose a new CIL method (VAG) that also leverages the  sparsity of vocabulary to focus the generation and creates pseudo-replay samples by using label semantics. Experimental results show that VAG outperforms baselines by a large margin.\footnote{Appeared in Proceedings of ACL 2023.}\footnote{Our code is publicly available at \url{https://github.com/shaoyijia/VAG}.} %

\end{abstract}

\section{Introduction}
\label{sec:intro}

Large pre-trained language models (PLMs) have become the \textit{de facto} standard in building NLP systems. However, how to best use them for continual learning (CL) is still a significant question~\citep{huang-etal-2021-continual,xia-etal-2022-learn,pasunuru-etal-2021-continual,ke-etal-2021-classic}. Many existing studies focus on \textit{task-incremental learning} (TIL) where the model learns distinct tasks sequentially and is given the task identity for inference. These works usually keep the PLM unchanged and update a series of additional structures such as adapters~\citep{gururangan-etal-2022-demix} or prompts~\citep{zhu-etal-2022-continual,qin2022lfpt}. Though effective, these methods cannot be used in a more challenging setting of \textbf{class-incremental learning} (CIL) which does not provide task information at test time.

CIL aims to build a single model to make predictions over incrementally learned classes organized as tasks %
(formal definition in~\cref{sec:background}). \citet{wu2022pretrained} conducted a comparative study on PLM in CL and showed that PLMs perform extremely poorly in the CIL setting due to \textit{catastrophic forgetting} (CF)\footnote{CF means that a neural network forgets previously learned knowledge when trained on new tasks, resulting in a decline in performance on earlier tasks~\citep{mccloskey1989catastrophic}.}. Also, as the task information is unknown, CIL further requires the model to predict the task identity of each test instance correctly.

\begin{figure}[t]
    \centering 
    \resizebox{\columnwidth}{!}{%
    \includegraphics{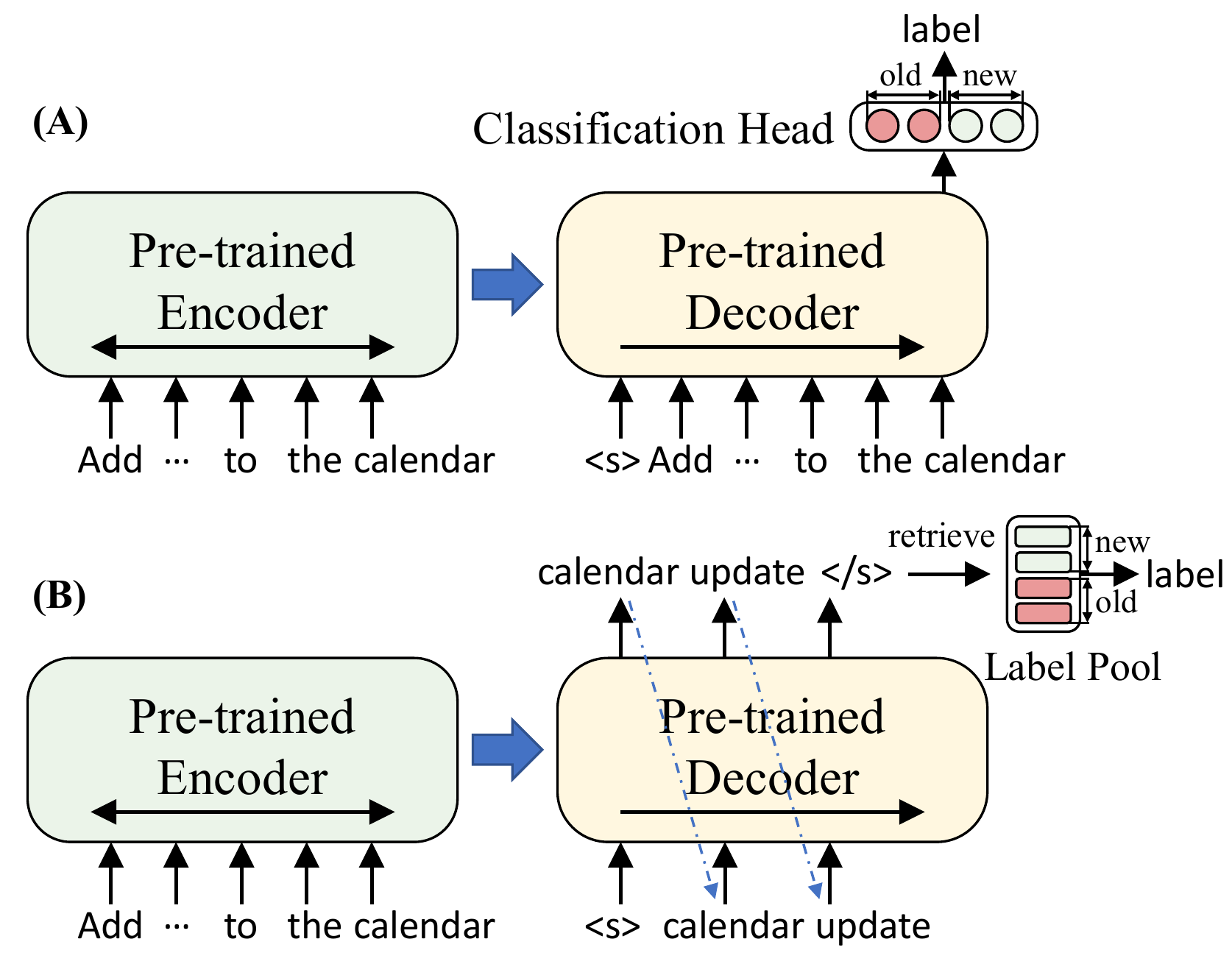}
    }
    \caption{Comparison between classifier framework (A) and generation framework (B) of using a pre-trained encoder-decoder model for class-incremental learning.}
    \label{Fig:intro}
\vspace{-0.5em}
\end{figure}

In this work, we re-examine the problem of using PLM for CIL and discovered that \textit{formulating CIL as \textbf{continual label generation} can greatly improve PLMs' continual learning ability}. As illustrated in~\reffig{Fig:intro}, a traditional classifier views the PLM as a large feature extractor and uses a linear classification head to map the extracted features to a probability distribution on both old and new labels. However, we can also use a generation approach to directly fine-tune the PLM to generate a label sequence (indicating a label) for a test instance. The final label is retrieved from the label pool of the classes learned so far based on text similarity. %

Some existing CL works have leveraged generation. For example, LAMOL~\citep{sun2019lamol} is a TIL system that uses generation to unify different types of tasks and creates pseudo replay samples; \citet{zhang-etal-2022-continual} focuses on the continual learning of different generation tasks.\footnote{Readers can refer to Appendix~\ref{sec:related_work} for more \textbf{related works}.} Different from these works, we are the first to directly use the generation objective to effectively ease the CF issue in the CIL process. Our experiments demonstrate that the generation objective is more suitable to the continual learning of PLM. To study the inner working of the paradigm shift, in~\cref{sec:generation}, we quantitatively show that the generation objective can prevent the PLM from representation collapse~\citep{aghajanyan2021better}, thus preserving its ability to continually learn new classes.

To further improve the generation approach, we propose the \textbf{VAG} (\textbf{V}ocabulary-\textbf{A}ware Label \textbf{G}eneration) system for CIL. %
VAG modifies the generation loss by focusing on different vocabulary subsets when learning different tasks. Owning to the natural sparsity of vocabulary, %
the modified loss leads to a sparse model update that greatly eases the CF issue. Moreover, VAG exploits the %
label semantics to create pseudo replay data via a label-based augmentation. Extensive experiments on 5 datasets show that VAG drastically outperforms baselines in non-exemplar based CIL (\ie, without saving any replay sample) and also achieves better results when a small amount of saved replay data is used.%

\section{Background}
\label{sec:background}
\textbf{Class-Incremental Learning (CIL).}\quad
CIL learns a sequence of tasks $\{1,...,T\}$ incrementally~\cite{kimtheoretical}. Each task $t$ learns a set of new classes $\mathcal{C}_t$. At task $t\in\{1,...,T\}$, 
the system is given a training set $\mathcal{D}_t = (\mathcal{X}_t, \mathcal{Y}_t)$, where $\mathcal{X}_t=\{x_j^{(t)}\}_{j=1}^{N_t}$ is the input data, $\mathcal{Y}_t=\{y_j^{(t)}\}_{j=1}^{N_t}$ is the set of their class labels and $y_{j}^{(t)} \in \mathcal{C}_t$. %
The classes in different tasks are disjoint, $\mathcal{C}_t \cap \mathcal{C}_{t^{\prime}}=\emptyset, \forall t^{\prime} \neq t$. %
At inference, given a test instance, the system selects a class label from $\bigcup_{t=1}^{T}\mathcal{C}_t$ \textit{without knowing the task identity}. The performance of the system is evaluated in the accuracy of the test samples from all seen classes.

\vspace{+1mm}
\noindent
\textbf{Encoder-Decoder Model}\quad
Encoder-decoder models take a sequence of tokens as input $X = x_1, ..., x_n$ and generate the target sequence $Y = y_1, ..., y_m$ in an auto-regressive manner. Specifically, the encoder maps the input sequence to a vector representation $z = f_{\theta_{enc}}(X)\in \mathbb{R}^{d_{enc}}$. Suppose the auto-regressive decoder has already generated $Y_{1:i-1}=y_1,...,y_{i-1}$, the next-token probability is 
\begin{equation}
    \resizebox{0.88\columnwidth}{!}{$%
    P(y_i|z, Y_{1:i-1}) = \frac{\operatorname{exp}(E_{y_i}^{\mathsf{T}}f_{\theta_{dec}}(z, Y_{1:i-1}))}{\sum_{w\in\mathcal{V}}\operatorname{exp}(E_{w}^{\mathsf{T}}f_{\theta_{dec}}(z, Y_{1:i-1}))}.
$}
\label{eq:decoder}
\end{equation}
Here, $E_{w}\in\mathbb{R}^{d_{dec}}$ denotes the word embedding of token $w\in \mathcal{V}$, where $\mathcal{V}$ is the model vocabulary. The model parameters are optimized to minimize the negative log-likelihood of ground truth $y_t$.%

\section{VAG System}
\label{sec:method}

\begin{figure}[t]
    \centering 
    \resizebox{\columnwidth}{!}{%
    \includegraphics{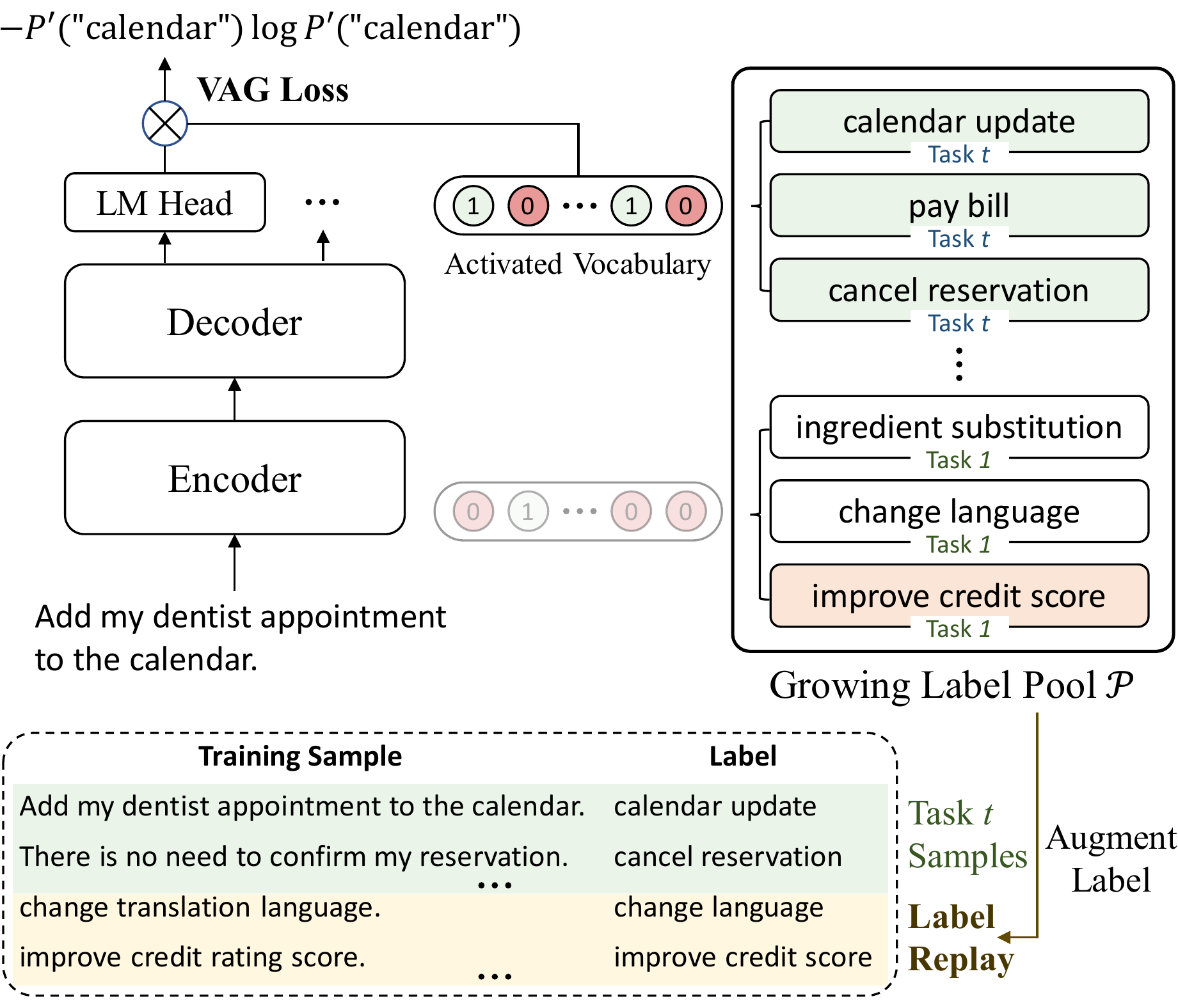}
    }
    \caption{Overview of training VAG on task $t$. VAG modifies the generation loss by masking the probability of unused vocabulary and creates pseudo replay data by augmenting the label sequences.}
    \label{Fig:method}
\vspace{-0.5em}
\end{figure}

We present the proposed VAG system which reframes CIL as a continual label generation problem. \reffig{Fig:method} gives an overview of VAG with two major components.

\subsection{Classification via Generation}
\label{sec:generation}
\begin{figure}[t]
    \centering 
    \resizebox{\columnwidth}{!}{%
    \includegraphics{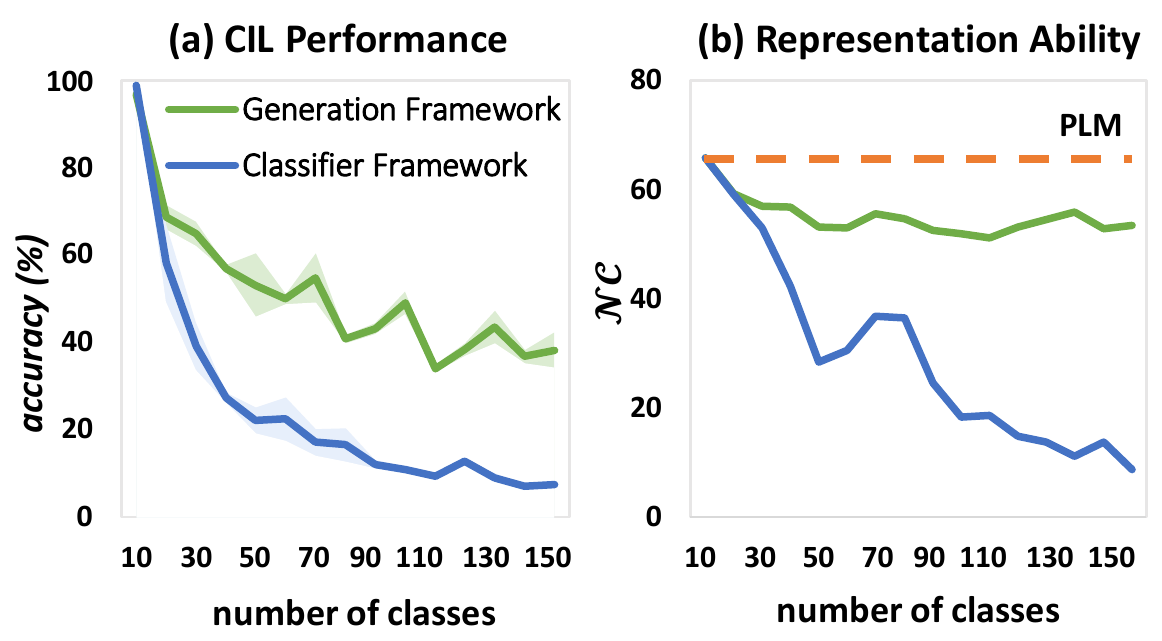}
    }
    \caption{Accuracy (\%) and $\mathcal{NC}$ (neural collapse) comparison of the classifier framework and generation framework for CIL on CLINC150 (15 tasks). %
    For both \textit{accuracy} and $\mathcal{NC}$, higher numbers are better.}
    \label{Fig:gen}
\vspace{-0.7em}
\end{figure}

VAG solves classification via label generation and maintains a label pool $\mathcal{P}$ of label sequences. Each label $c\in\mathcal{C}_t$ is a sequence of tokens representing a class label. When training task $t$, instead of mapping $\mathcal{C}_t$ to integer indexes representing class labels, VAG retains the label semantics and fine-tunes the PLM $\mathcal{M}$ to generate the label sequence conditioned on the input sequence $x_j^{(t)}$. In the CIL process, $\mathcal{P}$ keeps growing to contain all distinct label sequences seen so far. At inference, the most relevant label sequence will be retrieved from $\mathcal{P}$ based on the similarity between all the candidate labels and $y_{\text{gen}}$ generated by $\mathcal{M}$ given the input $x$:
\begin{equation}
\resizebox{0.88\columnwidth}{!}{$%
\begin{gathered}
y_{\text{gen}} = \operatorname{generate}(\mathcal{M},x)\\
y_{\text{pred}} = \argmax_{y\in\mathcal{P}} \cos({\operatorname{embed}(y), \operatorname{embed}(y_{\text{gen}})})
\end{gathered}$%
}
\label{equation:inference}
\end{equation}
Here, $\operatorname{embed}(\cdot)$ is parameterized by a Sentence-BERT model~\citep{reimers-gurevych-2019-sentence}.

Although the idea of solving CIL via generation is simple, the framework change yields a great performance boost. %
~\reffig{Fig:gen} compares the classifier framework and the generation framework on CLINC150~\citep{larson-etal-2019-evaluation} which contains 150 classes and is split into 15 tasks. With no additional mechanism to handle CF, using the same PLM, \ie $\text{BART}_{\text{base}}$~\cite{lewis-etal-2020-bart}, the generation framework gives much better results. 

\vspace{+1mm}
\noindent
\textbf{Generation loss prevents PLMs from collapsing.}\quad To understand the inner working of the framework change, we look into the PLM's representation ability in the CIL process. Unlike single-task learning, CIL requires the PLM to maintain the representation ability as much as possible for future classes, which is nontrivial because PLMs tend to have representation collapse\footnote{Representation collapse refers to the degradation of generalizable representations of pre-trained models during fine-tuning~\citep{aghajanyan2021better}.} during fine-tuning~\citep{aghajanyan2021better}. \reffig{Fig:gen} (b) %
compares the change of the PLM's representation ability in the two frameworks by using the neural collapse metric ($\mathcal{NC}$) proposed in~\citet{zhu2021geometric}:
\begin{equation}
\resizebox{0.52\columnwidth}{!}{$%
    \mathcal{NC}:=\frac{1}{K} \operatorname{trace}\left(\boldsymbol{\Sigma}_W \boldsymbol{\Sigma}_B^{\dagger}\right),
$}
\end{equation}
where $\boldsymbol{\Sigma}_W,\boldsymbol{\Sigma}_B\in\mathbb{R}^{d_{\text{enc}}\times d_{\text{enc}}}$ denote the within-class and between-class covariance matrices of the encoded sequences, $\boldsymbol{\Sigma}_B^{\dagger}$ denotes the pseudo inverse of $\boldsymbol{\Sigma}_B$, and $K$ denotes the number of classes in the dataset. As clearly shown, when learning more and more tasks, both frameworks witness a drop of the PLM's representation ability. However, the PLM in the generation framework keeps a relatively steady representation ability in the CIL process, thus remaining capable of learning unseen classes.

\subsection{Vocabulary-Aware Generation Loss}
\label{sec:loss}
One major challenge of CIL is that the previously learned decision boundaries may be corrupted when the model weights are updated to learn new classes~\citep{zhu2021class}. Beyond using the generation framework to retain the PLM's representation ability, we further propose a \textit{vocabulary-aware generation loss} (VAG loss) to ease the task interference (which causes catastrophic forgetting). 

Note that although the PLM is pre-trained with a large vocabulary (\eg, BART has a vocabulary size of 50,265), only a tiny subset will be used for the label generation in each task. VAG loss leverages this natural sparsity of vocabulary by masking the probability of tokens that will not be used in the current task before calculating the generation loss. Specifically, denote the vocabulary set of $\mathcal{C}_t$  as $\mathcal{V}_t$,  %
$P(y_i|z, Y_{1:i-1})$ in~\refequ{eq:decoder} is changed to
\begin{equation}
\resizebox{0.88\columnwidth}{!}{$%
    P'(y_i|z, Y_{1:i-1}) = \frac{\operatorname{exp}(E_{y_i}^{\mathsf{T}}f_{\theta_{dec}}(z, Y_{1:i-1}))}{\sum_{w\in\mathcal{V}_t}\operatorname{exp}(E_{w}^{\mathsf{T}}f_{\theta_{dec}}(z, Y_{1:i-1}))}.
$}
\label{eq:restrict}
\end{equation}
Since $|\mathcal{V}_t|\ll|\mathcal{V}|$, maximizing the modified probability leads to a sparse update of $E$ and effectively eases the forgetting of previous classes.

\subsection{Label-based Pseudo Replay}
\label{sec:replay}
Another major challenge of CIL is that the system needs to separate new classes in task $t$ and classes in previous tasks since the task identity is unknown at inference. %
To help construct decision boundaries across tasks and mitigate forgetting, VAG creates pseudo replay data by \textit{augmenting the label sequences} in previous tasks.

Specifically, given the label sequence $y$, the augmented sequence $\operatorname{aug}(y)$ will be used as a pseudo replay data instance with label $y$. To preserve the label semantics as well as to create diverse samples, we implement $\operatorname{aug}(\cdot)$ by randomly adding related tokens %
to the original label sequence based on contextual word embeddings~\citep{ma2019nlpaug}:
\begin{equation}
\resizebox{0.65\columnwidth}{!}{$%
    \mathcal{D}_{<t}^{LPR} = \{(\operatorname{aug}(y), y)|y\in\cup_{i=1}^{t-1}\mathcal{Y}_i\}
$}
\label{equ:label_aug}
\end{equation}
When training task $t$, we sample $\lambda|\mathcal{D}_t|$ pairs from $\mathcal{D}_{<t}^{LPR}$ ($\lambda$ is a hyper-parameter), %
and combine them with $\mathcal{D}_t$ as the training data. The VAG loss is also applied to the pseudo replay sample $(\operatorname{aug}(y), y)$, \ie, for each $y\in\mathcal{Y}_i$, its associated vocabulary subset $\mathcal{V}_i$ will be used in the denominator in~\refequ{eq:restrict}.

\section{Experiments}
\label{sec:experiment}

\begin{table*}[h]
\centering
\resizebox{\textwidth}{!}{%
\begin{tabular}{lccccclcccccccc|c} 
\toprule
                      & \multirow{2}{*}{\textbf{\#Tasks}} & \multicolumn{4}{c}{\textbf{Softmax Classifier}} &  & \multicolumn{9}{c}{\textbf{Generation}}                          \\ 
\cmidrule{3-6}\cmidrule{8-16}
                      & & Vanilla & EWC   & KD    & L2P                   &  & Vanilla-G & EWC-G   & KD-G    & L2P-G   & LAMOL & PAGeR & ACM   & \textbf{VAG} & {\cellcolor[rgb]{0.925,0.925,0.925}}Non-CL  \\ 
\midrule
\textbf{CLINC150}     & 15 & 7.37   & 7.67 & 9.39 & 3.32                 &  & 37.63   & 44.23 & 36.51 & 43.84 & 42.56 & 39.39 & 48.78 & \textbf{65.69}    & {\cellcolor[rgb]{0.925,0.925,0.925}}94.66      \\
\textbf{Banking77}    & 7 &  14.43       &  14.51     &     14.59  &    1.98       &  &     26.88    &  29.99     & 21.36      &    34.42   &  39.51  & 43.85   &  54.72     & \textbf{55.19}    & {\cellcolor[rgb]{0.925,0.925,0.925}}88.61           \\
\textbf{20News}       & 10 &  9.96       & 9.96      &     10.00  & 6.84                      &  &  44.17       & 49.81      & 30.84      &     25.47  &    52.05  & 49.61  &  60.79     & \textbf{73.51}    & {\cellcolor[rgb]{0.925,0.925,0.925}}86.81           \\
\textbf{FewRel}       & 8 &   12.39      &  13.09     &     12.33  &        6.60               &  &     19.44    & 25.12      & 15.95      &    6.52   &    34.69  &  39.09  &  29.74     &  \textbf{52.26}     & {\cellcolor[rgb]{0.925,0.925,0.925}}85.14         \\
\textbf{TACRED}       & 8 &     10.96    & 10.41      & 12.04      &    4.85                   &  &     23.44    &  24.36     & 17.44      &    10.18   & 16.46  &  27.99    & 18.67      & \textbf{46.15}       & {\cellcolor[rgb]{0.925,0.925,0.925}}70.38        \\
\hline
\textbf{\textit{Avg.}}       & \textbackslash{} &     11.02    & 11.13      & 11.67      &    4.72                   &  &     30.31    &  34.70     & 24.42      &    24.09   & 37.05  &  39.99    & 42.54      & \textbf{58.56}  & {\cellcolor[rgb]{0.925,0.925,0.925}}85.12\\
\bottomrule
\end{tabular}
}
\caption{Final accuracy (\%) of VAG and baseline methods for non-exemplar based CIL. The gray column shows the results in the non-continual learning setting which provides an upper bound. The reported results are averaged over 5 random seeds and the \textbf{standard deviations} are reported in Appendix~\ref{appendix:std}.%
}
\label{table:main}
\end{table*}

\subsection{Datasets and Baselines}
\label{sec:datasets_baselines}

\textbf{Datasets.}\quad
We use 5 datasets. Following \citet{wu2022pretrained}, we randomly split each dataset into X tasks with Y classes per task, expressed as (X/Y). 
CLINC150~\citep{larson-etal-2019-evaluation} (15/10) and Banking77~\citep{casanueva-etal-2020-efficient} (7/10) for intent classification, 20 Newsgroups (20News)~\citep{Lang95} (10/2) for topic classification, FewRel~\citep{han-etal-2018-fewrel} (8/10) and TACRED~\citep{zhang-etal-2017-position} (8/5) for relation classification. Additional details about the datasets are given in Appendix~\ref{appendix:dataset}.

\noindent
\textbf{Baselines.}\quad
We consider the following baselines: (1) \textbf{Vanilla} fine-tunes the PLM sequentially. (2) \textbf{EWC}~\citep{kirkpatrick2017overcoming} is a regularization-based method. (3) \textbf{KD}~\citep{hinton15distill} uses knowledge distillation. (4) \textbf{L2P}~\citep{wang2022learning} dynamically prompts the PLM without the task identity. These baselines use the classifier framework, and we adapt them to the generation framework as another set of baselines (\textbf{X-G}). %
We also consider 3 methods which use generation for CL: (5) \textbf{LAMOL}~\citep{sun2019lamol} fine-tunes GPT-2 continually with manual prompts and incorporates pseudo replay. Since LAMOL is a TIL system, we adapt it to CIL by using the same prompt. (6) \textbf{PAGeR}~\citep{varshney-etal-2022-prompt} extends LAMOL with contrastive training and knowledge distillation. (7) \textbf{ACM}~\citep{zhang-etal-2022-continual} extends LAMOL by adding compositional adapters. {ACM is not designed for classification, so we adapt it by training the PLM to generate the class label.
}%

\noindent
\textbf{Implementation details} are  in Appendix~\ref{appendix:implement}.

\subsection{Main Results}
\label{section:main_results}

\reftab{table:main} shows the results in the non-exemplar (non-replay) based CIL setting. The reported results are averaged over 5 random seeds.

\noindent
\textbf{Baselines using the generation objective give better results.}\quad
In accord with the findings in~\citet{wu2022pretrained}, regularization-based methods (\eg, EWC, KD) perform poorly. For L2P, although it keeps the PLM fixed, the algorithm cannot converge in our experiments due to the randomness introduced by the error-prone prompt selection. Comparing the same method in two frameworks (\eg, EWC \textit{v.s.} EWC-G), we can see that the framework switch is highly effective, which indicates the superiority of solving CIL via label generation. Moreover, the best-performing baseline ACM also adopts the generation objective.

\noindent
\textbf{Superiority of VAG.}\quad
On all the datasets, VAG achieves the best performance, even outperforming other baselines in the generation framework by a large margin (\reftab{table:main}). \reffig{Fig:ablation} also shows that VAG has less forgetting in the CIL process than the two best baselines. However, compared with the results in the non-continual learning setting (Non-CL in~\reftab{table:main}) which represent the performance upper bound for each dataset, our method still has considerable room for improvement, thereby encouraging future endeavors.

\noindent
\textbf{Extending VAG to use real replay data.}\quad
Notably, VAG can be directly extended to utilize real or saved replay data when they are available. Since real replay data are from the training distribution, we optimize the original generation loss upon the combination of $\mathcal{D}_t$ and the real replay data besides optimizing the VAG loss.\footnote{More details are included in Appendix~\ref{appendix:exemplar}.}
We consider \textbf{ER}~\citep{lopez2017gradient}, \textbf{DER++}~\citep{buzzega2020dark} and \textbf{LDBR}~\citep{huang-etal-2021-continual} as replay-based baselines and experiment with different replay buffer sizes. \reftab{table:replay} shows the comparison results. %
VAG still performs the best, especially when the buffer size is small (see the \textit{Avg.} row)\footnote{When the buffer size is large, all the methods approach the non-CL results (performance upper bound), so the performance gap between VAG and other baselines gets smaller.}.%

\subsection{Ablation Study and Analysis}
\begin{table*}[h]
\centering
\resizebox{\textwidth}{!}{%
\begin{tabular}{lccccccccccccccc} 
\toprule
                   & \multirow{2}{*}{\makecell[c]{\textbf{Ours}\\\textbf{(non-exemplar)}}} & \multicolumn{4}{c}{\textbf{Buffer size = 1\%}} &  & \multicolumn{4}{c}{\textbf{Buffer size = 3\% }} &  & \multicolumn{4}{c}{\textbf{Buffer size = 5\% }}  \\ 
\cmidrule{3-6}\cmidrule{8-11}\cmidrule{13-16}
                   &                       & ER    & DER++ & LDBR &  \textbf{VAG}                           &  & ER    & DER++ & LDBR & \textbf{VAG}                            &  & ER    & DER++ & LDBR & \textbf{VAG}                             \\ 
\midrule
\textbf{CLINC150}  & \cellcolor[rgb]{0.925,0.925,0.925}65.69                 & 55.62 & 56.85 & 67.34 & \textbf{72.44}                          &  & 78.06 & 73.29 & 81.34& \textbf{81.53}                           &  & 85.31 & 80.37 & \textbf{86.49} & 85.00                            \\
\textbf{Banking77} &  \cellcolor[rgb]{0.925,0.925,0.925}55.19                     &  45.24     &  48.32   & 54.76  &    \textbf{58.96}                            &  & 65.22      &  65.73  & 70.16 &  \textbf{70.57}                               &  & 74.32      &  73.06   & 74.37  & \textbf{74.81}                                 \\
\textbf{20News}    &   \cellcolor[rgb]{0.925,0.925,0.925}73.51                    &  84.53     & 84.24   & \textbf{85.30}   & 84.76                               &  &    85.45   & 85.30    & \textbf{86.53}  &  85.29                               &  &    85.79   & 85.66    & \textbf{86.83}  & 85.85                                 \\
\textbf{FewRel}    &   \cellcolor[rgb]{0.925,0.925,0.925}52.26                    & 60.77      & 63.21    & 51.26  & \textbf{68.56}                               &  &    74.20   & 72.92   & 65.21   &  \textbf{75.99}                               &  &    78.08   & 78.09   & 70.48   & \textbf{78.42}                                 \\
\textbf{TACRED}    &    {\cellcolor[rgb]{0.925,0.925,0.925}}46.15                   &  36.09     & 37.03 & 38.21      & \textbf{49.70}                               &  &     49.66  &  52.12     & 46.93 &  \textbf{58.00}                               &  &    56.93   & 55.72    & 52.22  & \textbf{61.28}                                 \\
\hline
\textbf{\textit{Avg.}}    &    {\cellcolor[rgb]{0.925,0.925,0.925}}58.56                   &  56.45     & 57.93 & 59.37      & \textbf{66.88}                               &  &     70.52  &  69.87     & 70.03 &  \textbf{74.28}                               &  &    76.09   & 74.58    & 74.08  & \textbf{77.07}                                 \\
\bottomrule
\end{tabular}
}
\caption{Final accuracy (\%) of VAG and exemplar-based baselines for CIL with different buffer sizes (\ie, we save 1\%, 3\%, 5\% of previous training data). The \textbf{standard deviations} are reported in Appendix~\ref{appendix:std}.}
\label{table:replay}
\end{table*}

\label{section:analysis}
\begin{figure}[t]
    \centering 
    \resizebox{\columnwidth}{!}{%
    \includegraphics{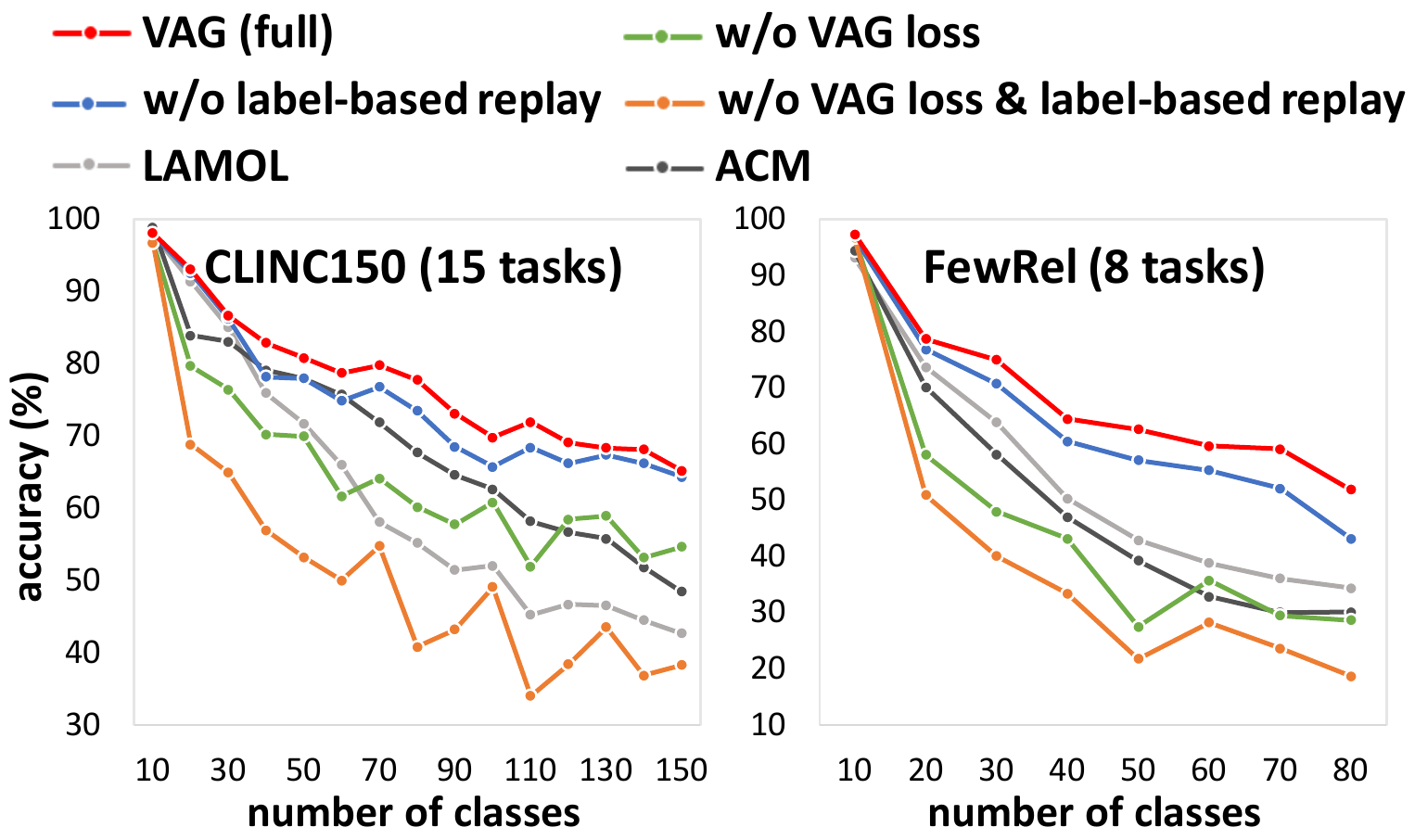}
    }
    \caption{Changes in accuracy (\%) with increasing tasks through the class-incremental learning process.}%
    \label{Fig:ablation}
\end{figure}

We analyze the effect of each component in our VAG system and~\reffig{Fig:ablation} shows the ablation results. While the full VAG uniformly gives the best results, we further observe that: (1) Both VAG loss and label-based replay can benefit CIL independently. (2) Label-based replay has a relatively small effect especially when we have already adopted VAG loss.

In Appendix~\ref{appendix:cm}, we compare the confusion matrices of ``VAG (full)'' and ``w/o VAG loss''. We find VAG loss effectively prevents the model from biasing towards predicting the latest learned classes, thus effectively easing the forgetting issue. In Appendix~\ref{appendix:label_replay}, we further analyze the impact of different label-based replay ratios ($\lambda$ in \cref{sec:replay}). \reffig{Fig:label_replay_ratio} shows that a small amount of label-based replay data already improves the results markedly, indicating the usefulness of leveraging label semantics for pseudo replay.

As discussed in~\cref{sec:generation}, the generation loss eases the drop of the PLM's representation power %
in the CIL process. Appendix~\ref{appendix:nc} reports the neural collapse metric $\mathcal{NC}$ of different methods after CIL. %
The VAG system preserves the representation ability of the PLM to the greatest extent.

\section{Conclusion}
We presented the VAG system which solves CIL based on label generation. We showed that migrating to the generation framework gives a drastic performance boost and eases the representation collapse of the pre-trained model. Experimental results demonstrate the effectiveness of VAG.

\section*{Limitations}
One limitation of this work is that VAG does not achieve zero forgetting. Although we show solving CIL based on label generation can effectively ease forgetting and representation collapse of the pre-trained model, it is still interesting to further explore how to explicitly solve the forgetting issue in this new framework. The proposed techniques in VAG are a step in the exploration.

Another limitation is that we directly use the label sequences provided by the original dataset. This may be suboptimal because the quality of the manually created label is hard to guarantee as it may fail to capture the semantic information of the samples in a class. A potential direction is to study creating label sequences automatically by summarizing the training samples. We leave this for future work.

\section*{Ethics Statement}
{While our proposed VAG system involves generation, it does not have the general ethical concern of generation, \ie, outputting biased or discriminative texts, because the final output of the system is retrieved from the label pool which is highly controllable. For our experiments, we use public datasets and believe that none of them contains offensive contents. Also, although the training of the VAG system requires computational resources, the CIL paradigm is resource-efficient because the model preserves the previously learned knowledge and continually learns new classes.
}

\bibliography{anthology,custom}
\bibliographystyle{acl_natbib}

\clearpage
\appendix

\label{sec:appendix}
\section{Related Work}
\label{sec:related_work}
\textbf{Continual Learning.}\quad
Continual learning requires a model to sequentially learn a series of tasks. The main challenge that existing papers focus on is overcoming \textit{catastrophic forgetting} (CF)~\citep{mccloskey1989catastrophic}. Previous works usually fall in the following categories: (1) Regularization-based methods, which penalize the parameter update and preserve the previous task knowledge~\citep{kirkpatrick2017overcoming,huang-etal-2021-continual,zhu2021prototype,li2017learning}. (2) Parameter-isolation methods, which separate parameters for different tasks by finding subnetworks in the over-parameterized model~\citep{wortsman2020supermasks,serra2018overcoming,mallya2018packnet} or adding additional task-specific modules~\citep{houlsby2019parameter,ke-etal-2021-classic}. These methods need to know the task identity for inference. (3) Replay-based methods, which jointly train the model with new task data and some saved examples \citep{lopez2017gradient,buzzega2020dark} or generated pseudo data \citep{shin2017continual,sun2019lamol} of previous tasks. In real applications, storing replay samples may not be possible due to the data privacy issue or memory overhead~\citep{zhu2021prototype}.

Based on the differences in evaluation protocols, continual learning can be summarized into three major settings: class-incremental learning (CIL), task-incremental learning (TIL), and domain-incremental learning (DIL)~\citep{yin-etal-2022-contintin}. Among them, CIL which aims to build a single predictive model on all seen classes, is the most difficult one because the task identity is not available for inference. This requires the model to not only tackle catastrophic forgetting of the within-task prediction ability but also predict the task identity correctly~\citep{kimtheoretical}. In the language domain, prior works %
have studied CIL for intent detection~\citep{liu2021lifelong,li-etal-2022-continual}, relation classification~\citep{han-etal-2020-continual,zhao-etal-2022-consistent}, named entity recognition~\citep{monaikul2021continual,xia-etal-2022-learn}, \etc Despite the great success of pre-trained language models (PLMs), these models still suffer from severe CF issue in continual learning. In a large-scale comparative study, \citet{wu2022pretrained} concluded that PLMs perform extremely poorly in the CIL setting. In their study, a PLM is leveraged by fine-tuning the model with a classification head. However, in this work, we find that PLMs can show better CIL ability if we fine-tune the PLM in a generation framework.

\noindent
\textbf{Text Generation in Continual Learning Study.}\quad
With the success of natural language generation using PLMs~\citep{radford2019language, lewis-etal-2020-bart,raffel2020exploring}, some works on continual learning of NLP utilize the generation ability of PLMs to unify different potential tasks through prompting~\citep{qin2022lfpt} or instruction tuning~\citep{yin-etal-2022-contintin, scialom2022finetuned}. The text generation can also be used to create pseudo replay data for previous task. 
LAMOL~\citep{sun2019lamol} is a typical system in this line of work which simultaneously learns to solve all the tasks in a unified question-answering manner and generates pseudo replay samples in the TIL setting. While LAMOL is closely related to our work which also leverages generation, the key difference is that we focus on CIL instead of TIL and show for the first time that the generation objective itself can effectively ease the CF issue. We also show that the generation objective bears a link with preventing the representation collapse of the PLM and further propose the VAG approach to exploit the generation framework for CIL. 
Some other works in the continual learning literature directly focus on generation tasks (not classification tasks) and study the problem of continual sequence generation~\citep{zhang-etal-2022-continual,mi-etal-2020-continual}. These works naturally involve generation due to the property of their studied tasks.

\section{Additional Details of Experiments}
\label{appendix:exp_details}

\subsection{Dataset Details}
\label{appendix:dataset}
\begin{table}
\resizebox{\columnwidth}{!}{%
\centering
\begin{tabular}{llllll} 
\toprule
\textbf{Dataset}       & \textbf{Class} & \textbf{Task}                & \textbf{Train}  & \textbf{Validation}    & \textbf{Test}    \\ 
\midrule
CLINC150      & 150   & 15 & 15,000 & 3,000  & 4,500   \\
Banking77     & 77    & 7  & 7,191  & 1,800  & 2,800   \\
20News & 20    & 10  & 10,000 & 3,998  & 5,999   \\
FewRel        & 80    & 8  & 33,600 & 11,200 & 11,200  \\
TACRED        & 42    & 8   & 5,909  & 1,482  & 1,259   \\
\bottomrule
\end{tabular}
}
\caption{Dataset statistics. Banking77 and TACRED do not have the validation set, so we randomly sample 20\% data from the training set for validation.}
\label{table:statistics}
\end{table}
As described in~\cref{sec:datasets_baselines}, we use 5 datasets for our experiments. \textbf{CLINC150}~\citep{larson-etal-2019-evaluation} and \textbf{Banking77}~\citep{casanueva-etal-2020-efficient} are two intent classification datasets with 150 classes and 77 classes respectively. Each intent class is described by a short phrase (\eg, ``change language'', ``edit personal details'') in the original dataset, and we directly use these phrases as the label sequences. \textbf{20 Newsgroups (20News)} is a topic classification dataset with 20 categories associated with hierarchical labels (\eg, ``comp.sys.ibm.pc.hardware'' and ``misc.forsale''). We convert the hierarchical labels into label sequences by replacing ``.'' with a whitespace and extending the abbreviations into complete words (\eg, ``computer system ibm pc hardware'', ``miscellaneous forsale''). \textbf{FewRel}~\citep{han-etal-2018-fewrel} is a relation classification dataset with 80 relations. \textbf{TACRED}~\citep{zhang-etal-2017-position} is another relation classification dataset with 42 relations and it has highly unbalanced samples for each relation. In these two datasets, each relation is described by a short phrase (\eg, ``exhibition history'', ``organization related: founded by'') and we use them as the label sequences.

Following~\citet{wu2022pretrained}, we randomly split CLINC150, Banking77, FewRel into disjoint tasks with 10 classes per task. We split 20News into 10 tasks with 2 classes per task and TACRED into 8 tasks with 5 classes per task for a more challenging evaluation. \reftab{table:statistics} summarizes the dataset statistics.

Note that among the datasets we used, CLINC150\footnote{\url{https://github.com/clinc/oos-eval/blob/master/LICENSE}}, Banking77\footnote{\url{https://github.com/PolyAI-LDN/task-specific-datasets/blob/master/LICENSE}}, FewRel\footnote{\url{https://github.com/thunlp/FewRel/blob/master/LICENSE}}, TACRED\footnote{\url{https://catalog.ldc.upenn.edu/LDC2018T24}} are licensed. We ensure that we did not violate any license condition when conducting our experiments.

\subsection{Implementation Details}
\label{appendix:implement}
\begin{table*}
\resizebox{\textwidth}{!}{%
\centering
\begin{tabular}{l|ll|l} 
\toprule
\textbf{Method} & \textbf{Key} & \textbf{Value} & \textbf{Note}                                                                                           \\ 
\midrule
EWC             & $\lambda$       & 5,000          & The weight for penalty, selected from [500, 1,000, 2,000, 5,000, 10,000, 20,000, 50,000].               \\ 
\midrule[0.5pt]
KD              & $\lambda$      & 0.1            & The weight for knowledge distillation loss, selected from [0.1, 0.5, 1.0].                              \\ 
\midrule[0.5pt]
L2P             & $M$            & 10             & The total number of prompts, following the original paper.                                              \\
                & $N$            & 5              & The number of dynamically selected prompts, following the original paper.                               \\
                & $\lambda$       & 0.5            & The weight of key selection loss, following the original paper.                                         \\ 
\midrule[0.5pt]
LAMOL           & $\gamma$        & 0.2            & The sampling ratio of pseudo replay data, following the original paper.                                 \\ 
\midrule[0.5pt]
PAGeR           & $\lambda_1$      & 1              & The weight of the generation loss and distillation loss, following the original paper.                  \\
                & $\lambda_2$      & 0.25           & The weight of the replay data generation loss, following the original paper.                            \\
                & $\lambda_3$      & 0.25           & The weight of the supervised contrastive training loss, following the original paper.                   \\
                & $\gamma$        & 0.2            & Refer to $\gamma$ in LAMOL.                                                                                 \\ 
\midrule[0.5pt]
ACM             & $\gamma$        & 0.01           & The entropy coefficient, using the default value of the official implementation.                        \\
                & $c$            & 0.15           & The initialization of the coefficient weights, using the default value of the official implementation.  \\
\bottomrule
\end{tabular}
}
\caption{The hyper-parameters of baseline implementation.}
\label{table:implementation}
\end{table*}

We implement VAG and baseline (1)-(4) with the Transformers library~\citep{wolf-etal-2020-transformers} and use $\text{BART}_\text{base}$\footnote{\url{https://huggingface.co/facebook/bart-base}} (\#parameters: 139M) as the backbone PLM. For LAMOL\footnote{\url{https://github.com/chho33/LAMOL}} and ACM\footnote{\url{https://github.com/SALT-NLP/Adaptive-Compositional-Modules}}, we directly use their official implementation and use the same question prompt for each task\footnote{For CLINC150, Banking77, 20News, we set the question prompt to be ``What's the category of this text?''. For FewRel and TACRED, we set the question prompt to be ``What's the relation between these two entities?''.} so that they do not need the task identity for inference any more and can suit the CIL setting. %
For PAGeR, we use our own implementation because its source code is not publicly available. \reftab{table:implementation} gives the hyper-parameters of baseline implementations.

For learning each task, we train the model for 10 epochs and use the validation set of the current task for early stopping. We set the batch size as 8 and the max sequence length as 128. We use AdamW optimizer~\citep{loshchilov2018decoupled} with $\beta_1=0.9$, $\beta_2=0.999$ and the learning rate of 1e-5. For the label-based pseudo replay component of VAG, we implement $\operatorname{aug}(\cdot)$ using the \texttt{ContextualWordEmbdsAug} in the nlpaug library\footnote{\url{https://pypi.org/project/nlpaug/}} which adds $0.3\times\operatorname{token\_num}(y)$ related tokens to the original label sequence $y$ and the hyper-parameter $\lambda$ is set to 0.1. At inference, we use greedy decoding to decode the generated sequence and $\operatorname{embed}(\cdot)$ in \refequ{equation:inference} is parameterized by \texttt{paraphrase-MiniLM-L6-v2} provided in the Sentence-Transformers library\footnote{\url{huggingface.co/sentence-transformers/paraphrase-MiniLM-L6-v2}}. We use NVIDIA GeForce RTX 2080 Ti GPU to conduct all our experiments.

\subsection{Exemplar-Based Setting}
\label{appendix:exemplar}
As discussed in~\cref{section:main_results}, we extend VAG system to the exemplar-based CIL setting where real replay data are available. In exemplar-based CIL, the training objective of VAG at task $t$ is to minimize
\begin{equation}
\resizebox{\columnwidth}{!}{$%
    \mathbb{E}_{\mathcal{D}_{<t}^{ER}\cup\mathcal{D}_t}[\ell_{normal}(x, y)] +\mu\mathbb{E}_{\mathcal{D}_{<t}^{LPR}\cup\mathcal{D}_t}[\ell_{VAG}(x, y)],
$}
\label{equation:replay}
\end{equation}
where $\mathcal{D}_{<t}^{ER}$ represents the real replay data of previous tasks, $\mathcal{D}_{<t}^{LPR}$ represents the label-based pseudo replay data (see~\refequ{equ:label_aug}), and $\mu$ is a hyper-parameter balancing two replay terms. We set $\mu$ to 1 in our experiments.

For comparison, we consider 3 typical replay-based methods: (1) \textbf{ER}~\citep{lopez2017gradient} directly combines replay samples and current task samples in training batches to fine-tune the classifier. (2) \textbf{DER++}~\citep{buzzega2020dark} exploits replay data in training and adds a regularization term to prevent the logits of replay data from changing. (3) \textbf{LDBR}~\citep{huang-etal-2021-continual} uses information disentanglement based regularization and selects replay samples through K-means clustering. We experiment with different buffer sizes by storing 1\%, 3\%, and 5\% of previous training data. Other training hyper-parameters are in accord with the non-exemplar based setting.

\subsection{Standard Deviations}
\label{appendix:std}

\begin{table*}
\centering
\resizebox{\textwidth}{!}{%
\begin{tabular}{lcccclcccccccc|c} 
\toprule
                      & \multicolumn{4}{c}{\textbf{Softmax Classifier}} &  & \multicolumn{8}{c}{\textbf{Generation}}                          \\ 
\cmidrule{2-5}\cmidrule{7-15}
                      & Vanilla & EWC   & KD    & L2P                   &  & Vanilla-G & EWC-G   & KD-G    & L2P-G   & LAMOL & PAGeR & ACM   & \textbf{VAG} & Non-CL  \\ 
\midrule
\textbf{CLINC150}     & $\pm0.56$   & $\pm0.50$ & $\pm1.50$ & $\pm0.34$                 &  & $\pm2.95$   & $\pm1.72$ & $\pm1.44$ & $\pm4.99$ & $\pm0.74$ & $\pm3.04$ & $\pm2.50$ & $\pm1.54$ & $\pm0.67$          \\
\textbf{Banking77}    &  $\pm0.68$       &  $\pm0.51$     &     $\pm0.46$  &    $\pm0.40$       &  &     $\pm3.28$    &  $\pm2.02$     & $\pm0.83$      &    $\pm3.01$   &  $\pm0.92$  & $\pm2.78$   &  $\pm1.54$     & $\pm0.37$  & $\pm0.94$             \\
\textbf{20News} &  $\pm0.02$       & $\pm0.01$      &     $\pm0.04$  & $\pm0.35$                      &  &  $\pm3.43$       & $\pm5.04$      & $\pm2.02$      &     $\pm1.69$  &    $\pm2.80$ & $\pm1.55$  &  $\pm2.55$     & $\pm3.81$ & $\pm0.35$               \\
\textbf{FewRel}       &  $\pm0.30$      &  $\pm0.55$     &     $\pm1.06$  &        $\pm0.68$               &  &     $\pm1.26$    & $\pm1.14$      & $\pm1.13$      &    $\pm3.43$   &    $\pm1.41$ & $\pm1.69$  &  $\pm1.88$     &  $\pm1.29$   & $\pm0.73$           \\
\textbf{TACRED}       &     $\pm1.09$    & $\pm0.29$      & $\pm1.33$      &    $\pm0.30$                   &  &     $\pm1.08$    &  $\pm1.36$     & $\pm1.30$      &    $\pm0.94$   & $\pm0.26$  & $\pm1.08$    & $\pm1.76$      & $\pm0.59$   & $\pm0.33$            \\
\bottomrule
\end{tabular}
}
\caption{Standard deviations of the proposed VAG system and the baselines in non-exemplar based class-incremental learning setting. The corresponding averaged results are in~\reftab{table:main}.}
\label{table:main_std}
\end{table*}

\begin{table*}
\centering
\resizebox{\textwidth}{!}{%
\begin{tabular}{lccccccccccccccc} 
\toprule
                   & \multirow{2}{*}{\makecell[c]{\textbf{VAG}\\\textbf{(non-exemplar)}}} & \multicolumn{4}{c}{\textbf{Buffer size = 1\%}} &  & \multicolumn{4}{c}{\textbf{Buffer size = 3\% }} &  & \multicolumn{4}{c}{\textbf{Buffer size = 5\% }}  \\ 
\cmidrule{3-6}\cmidrule{8-11}\cmidrule{13-16}
                   &                       & ER    & DER++ & LDBR & \textbf{VAG}                           &  & ER    & DER++ & LDBR & \textbf{VAG}                            &  & ER    & DER++ & LDBR & \textbf{VAG}                             \\ 
\midrule
\textbf{CLINC150}  & $\pm1.54$                 & $\pm8.42$ & $\pm7.90$  & $\pm1.75$ & $\pm0.56$                          &  & $\pm3.35$ & $\pm0.83$ & $\pm1.52$ & $\pm0.88$                           &  & $\pm1.40$ & $\pm1.17$ & $\pm0.50$ & $\pm1.05$                            \\
\textbf{Banking77} &  $\pm0.37$                  &  $\pm6.38$     &  $\pm2.78$   & $\pm1.80$  &    $\pm1.95$                            &  & $\pm2.24$      &  $\pm1.21$  & $\pm0.09$ & $\pm1.72$                               &  & $\pm2.77$      &  $\pm1.69$  & $\pm2.48$   & $\pm1.18$                                 \\
\textbf{20News}    &   $\pm3.81$                    &  $\pm1.01$     & $\pm1.41$  & $\pm0.04$    & $\pm0.39$                               &  &    $\pm0.28$   & $\pm0.28$   & $\pm0.35$   &  $\pm0.49$                               &  &    $\pm0.28$   & $\pm0.07$ & $\pm0.34$     & $\pm0.28$                                 \\
\textbf{FewRel}    &   $\pm1.29$                    & $\pm3.37$      & $\pm4.91$     & $\pm1.46$ & $\pm0.94$                               &  &    $\pm0.92$   & $\pm1.41$   & $\pm1.41$   &  $\pm0.65$                               &  &    $\pm0.72$   & $\pm1.21$ & $\pm1.74$     & $\pm0.63$                                 \\
\textbf{TACRED}    &      $\pm0.59$                 &  $\pm3.85$     & $\pm3.97$   & $\pm0.71$   & $\pm2.02$                               &  &     $\pm2.61$  &  $\pm4.30$  & $\pm1.47$   &  $\pm3.24$                               &  &    $\pm2.96$   & $\pm1.75$ & $\pm1.43$     & $\pm0.99$                                 \\
\bottomrule
\end{tabular}
}
\caption{Standard deviations of the proposed VAG system and the baselines for class-incremental learning setting with different buffer sizes. The corresponding averaged results are in~\reftab{table:replay}.}
\label{table:replay_std}
\end{table*}

In~\cref{section:main_results}, we evaluated our proposed system VAG in both non-exemplar and exemplar-based CIL setting. \reftab{table:main_std} and \reftab{table:replay_std} give the standard deviations of the reported results.

\section{Confusion Matrices}
\label{appendix:cm}
\begin{figure}[t]
    \centering 
    \resizebox{\columnwidth}{!}{%
    \includegraphics{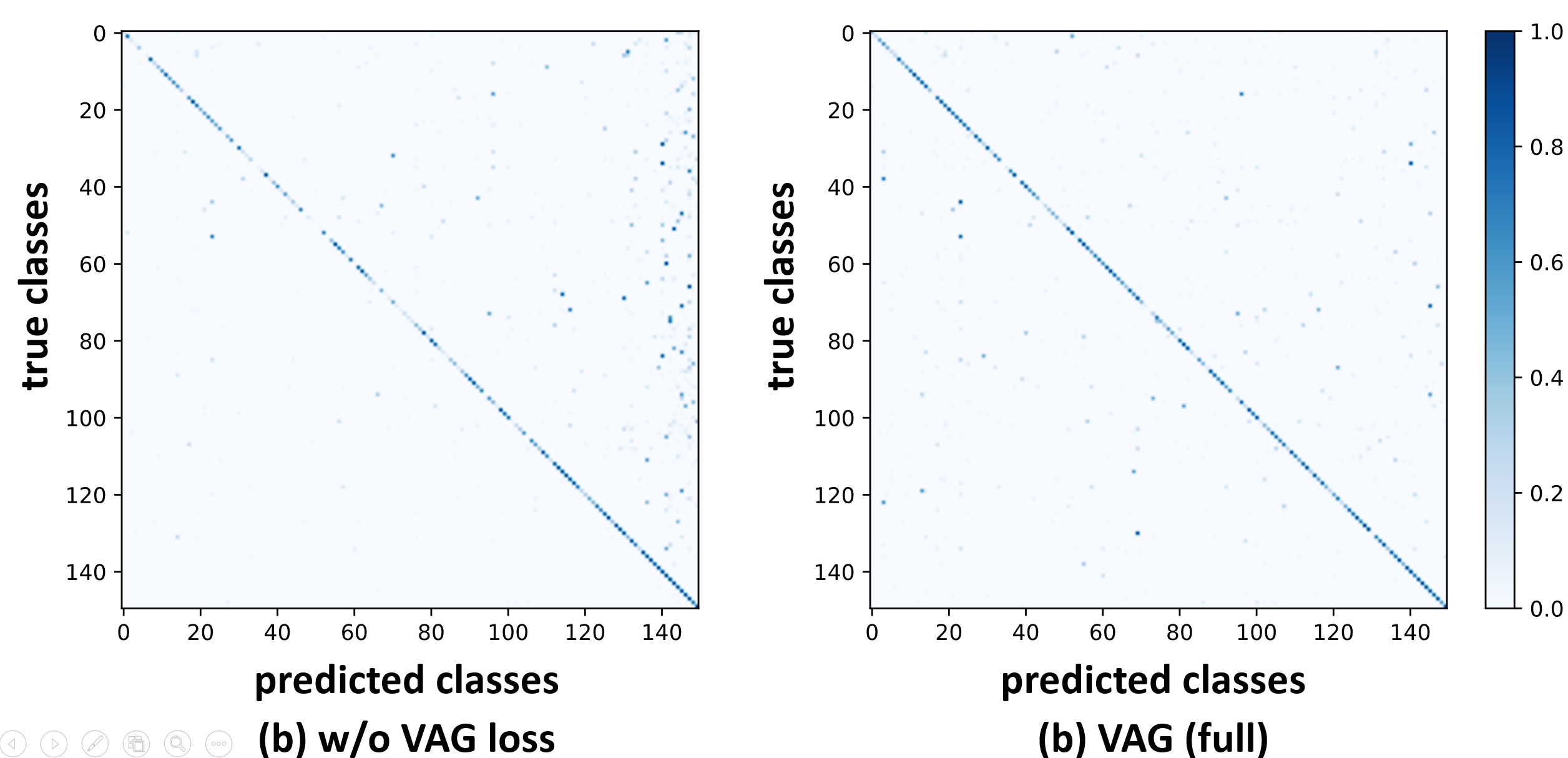}
    }
    \caption{Confusion matrix of ``VAG (full)'' and ``w/o VAG loss'' on CLINC150 (15 tasks).}
    \label{Fig:cm}
\end{figure}

In~\cref{section:analysis}, we analyze the effectiveness of each component in the proposed VAG system.  To study the  effect of VAG loss, we compare the confusion matrixes of ``VAG (full)'' and ``w/o VAG loss''. As shown in~\reffig{Fig:cm}, VAG loss effectively prevents the model from having a strong bias towards predicting the latest learned classes. Since VAG loss limits the denominator to the vocabulary used by the current task, training with VAG loss has less interference to previous task knowledge, thus yielding better final performance.

\section{Analysis of Label-Based Replay Ratio}
\label{appendix:label_replay}
\begin{figure}[t]
    \centering 
    \resizebox{0.75\columnwidth}{!}{%
    \includegraphics{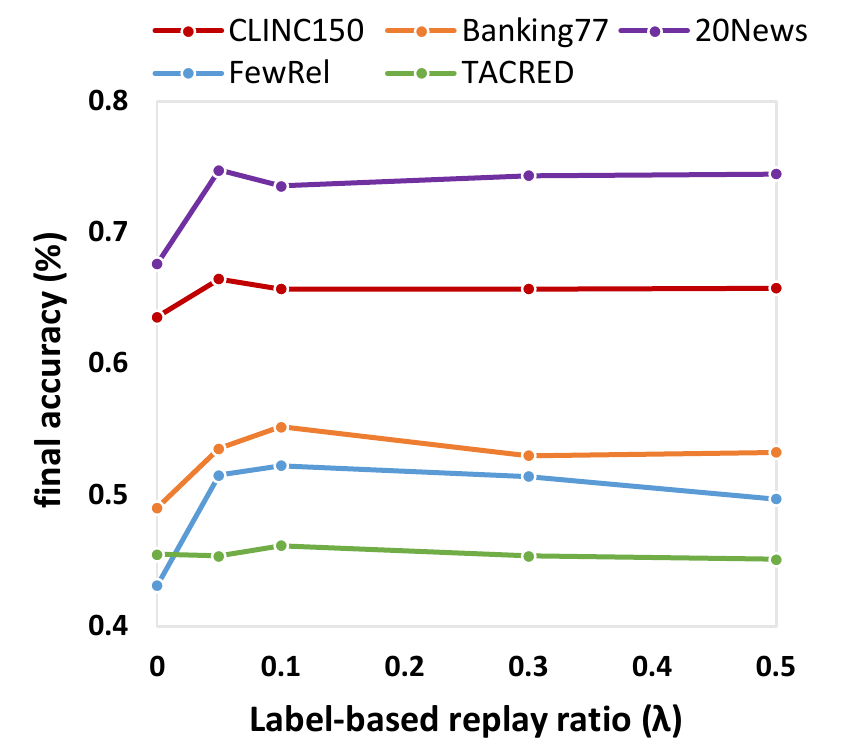}
    }
    \caption{Final results (accuracy) with different label-based replay ratios.}
    \label{Fig:label_replay_ratio}
\end{figure}
{
As discussed in~\cref{sec:replay}, VAG samples $\lambda|\mathcal{D}_t|$ pseudo replay data instances created by label-based data augmentation and combines them with $\mathcal{D}_t$ as the training data. Here, we analyze the impact of different label-based replay ratios $\lambda$.  \reffig{Fig:label_replay_ratio} shows the results. We observe that a small amount of label-based replay data can already yield improvements and the results are similar when we further increase the label-based replay ratio $\lambda$. We set $\lambda$ to 0.1 in our main experiments (see~\cref{sec:experiment}).
}
\section{Neural Collapse with Different Methods}
\label{appendix:nc}
As discussed in~\cref{sec:generation}, we find the generation framework can better preserve the representation ability of the pre-trained model in the CIL process. \reftab{table:nc} gives the neural collapse metric $\mathcal{NC}$ of different methods after CIL. In general, after the continual learning process, all the models have lower $\mathcal{NC}$ compared with the original PLM, especially when we fine-tuned the PLM using the traditional classifier framework. We also observe that while we modify the generation loss in the VAG system, its desired property is retained and our proposed CIL framework preserves the representation ability of the PLM to the greatest extent.
\begin{table}
\centering
\resizebox{\columnwidth}{!}{%
\begin{tabular}{lcccc} 
\toprule
                   & \makecell[c]{PLM\\(before CIL)}                            & Vanilla & Vanilla-G     & \textbf{VAG}     \\ 
\midrule
\textbf{CLINC150}  & {\cellcolor[rgb]{0.925,0.925,0.925}}65.84  & 8.70              & 53.47          & \textbf{57.24}   \\
\textbf{Banking77} & {\cellcolor[rgb]{0.925,0.925,0.925}}109.55 & 46.34             & \textbf{72.34} & 71.04            \\
\textbf{20News}    & {\cellcolor[rgb]{0.925,0.925,0.925}}15.92  & 2.16              & 13.95          & \textbf{15.51}   \\
\textbf{FewRel}    & {\cellcolor[rgb]{0.925,0.925,0.925}}321.09 & 77.31             & 170.25         & \textbf{190.09}  \\
\textbf{TACRED}    & {\cellcolor[rgb]{0.925,0.925,0.925}}46.79  & 32.78             & 40.54          & \textbf{45.54}   \\
\bottomrule
\end{tabular}
}
\caption{$\mathcal{NC}$ of models before and after class-incremental learning with different training methods.}
\label{table:nc}
\end{table}

\end{document}